\documentclass{knac}

\usepackage{hyperref}
\usepackage{float}

\begin{document}
\title{Node Classification and Search on the Rubik's Cube Graph with GNNs}


\author{Alessandro Barro\\
\scriptsize\texttt{alessandro1.barro@mail.polimi.it}
}

\begin{abstract}
\noindent This study focuses on the application of deep geometric models to solve the 3x3x3 Rubik's Cube. We begin by discussing the cube's graph representation and defining distance as the model's optimization objective. The distance approximation task is reformulated as a node classification problem, effectively addressed using Graph Neural Networks (GNNs). After training the model on a random subgraph, the predicted classes are used to construct a heuristic for $A^*$ search. We conclude with experiments comparing our heuristic to that of the DeepCubeA model.
\end{abstract}

%
%
\section{Introduction}
\noindent In this paper, we demonstrate how Deep Geometric Models can be applied to solve the 3x3x3 Rubik's Cube. Specifically, building on the approach of DeepCubeA [\cite{Agostinelli2019}], we employ a model to learn a heuristic leveraged by the $A^*$ search algorithm, while minimizing the role of human knowledge [\cite{McAleer2018}] by collecting training data through random walks. Unlike previous research [\cite{Yao2024}], which focuses on learning the graph’s local structure by identifying a vector of most probable actions and defining the Weighted Convolutional Distance, we reformulate the distance approximation task as a node classification problem. This framework can then be effectively addressed using Graph Neural Networks (GNNs). The label predictions from the GNN are then directly utilized to construct a consistent heuristic that guides the search on the graph. Initially, we formalize key concepts such as the Rubik's Cube Cayley graph and the notion of relative distance, which serve as foundations for the proposed methods. Subsequently, we describe how the problem is formulated as a classification task, as well as discussing about the model application. Finally, we present experimental results, make observations, and draw conclusions regarding the effectiveness and implications of this approach.
\section{Rubik's Cube Graph Representation}
\noindent Let us define the Rubik's Cube group $(G,\cdot)$, where $\cdot$ denotes the composition of permutations [\cite{Joyner2002}]. We introduce the generating set $S=\{U,D,L,R,F,B\}$, for which every element $g\in G$ can be expressed as a finite product of elements of $S$ and their inverses. Now, we define the associated Cayley graph $\Gamma(G,S,g_0)$ [\cite{Cayley1878}] as follows:
\begin{itemize}
    \item Nodes are the elements $g\in G$ and the graph is rooted in $g_0$, which reasonably indicates the solved configuration.
    \item Edges are valid transitions, under the application of the generators $s\in S$ and their inverses, between the group elements.
    \item Each edge is assigned a color $c_s$ to distinguish the moves that generate the transitions.
\end{itemize}
For practical purposes, we consider $\Gamma$ to be undirected, as each generator $s$ has an inverse $s^{-1}$ ensuring bidirectional transitions between nodes. This allows us to evaluate paths to $g_0$ from other, unsolved cube configurations. The order of $\Gamma$ is equal to the cardinality of $G$
\begin{equation}
    |G|\approx 4.3 \cdot 10^{19}
\end{equation}
which renders its complete computation out of the question. Notably, the graph shows high regularity in connectivity since $\text{deg}(g)=12$ $\forall g\in G$, although we will distinguish different node types in terms of their distance from the solution $g_0$ in the succeeding paragraphs. 

\section{Definition and Properties of the Distance}
\noindent Our analysis focuses on a distance metric defined between the elements of $G$ and the solution $g_0$. This metric forms the basis of our method, which ultimately employs a Deep Geometric Model as a heuristic for a local search task. Given \( g, g' \in G \), let \( d(g, g') \) denote the length of the shortest sequence of generators (and their inverses) in $S$ that transforms $g$ into $g'$. 

\noindent An important result [\cite{Rokicki2014}] shows that $0 \leq d(g,g') \leq M$, where $M = 26$ is the diameter of $G$ under the generating set $S \cup S^{-1}$. This implies that every cube configuration can be solved in at most $M$ moves. In addition, it is easy to show that the function $d : G \times G \to [0, M]$ satisfies the distance axioms in $G$, thus, $d$ is a valid distance metric on $G$. 

\noindent The primary object of interest in our analysis is not $d$ itself, but rather $d_0(g) = d(g_0, g) $, which measures the distance of an element $g$ relative to the solved cube state $g_0$. The latter metric enables further exploration of the Cayley graph's structure, particularly in terms of distance-based partitions and their implications for search and optimization algorithms.

\section{Node Classification for Distance Estimation}
\noindent The previously defined relative distance $d_0$ permits to identify $M$ distinct level structures of $\Gamma$ [\cite{Diaz2002}], leading to a distance-based partition of the set of vertices $\mathcal{Y}_n = \{ g \in G : d_0(g) = n \}, \quad n = 0, 1, \dots, M$. Such subsets have significant implications for the transition dynamics of the cube and make us able to distinguish various types of nodes. Each node has exactly 12 edges, corresponding to the possible moves in $S \cup S^{-1}$ and, importantly, any move either increases or decreases $d_0(g)$. Distinguishing nodes by the number $k \in [0,12]$ of edges that decrease (or increase) the distance is critical in the context of random walks, which we will employ, as well as in path-solving strategies, since in both cases probabilities are associated with each move. 

\noindent While it trivially holds that $d_0(s \cdot g_0) = 1$ and $d_0(s \cdot g_M) = M-1$ for all moves $s$, the composition of $k$-nodes within $\Gamma$ 
heavily influences the distribution of the elements $g$ with respect to $d_0$ in the Cayley graph.
To treat the latter distribution, we define a random variable $Y_g$ associated to $g$, which takes values in $\{0, 1, \dots, M\}$, corresponding to the level structure $\mathcal{Y}_n$ to which $g$ belongs. This  framework allows us to assign each node $g$ to a class $\mathcal{Y}_n$ according to the probability $p(g)=P(Y_g=y_g)$. Thus, formally reconducting the distance approximation problem to a node classification task, which can be tackled by Deep Geometric Models such as GNNs.

\begin{figure}[H]
    \centering
    \includegraphics[width=1\linewidth]{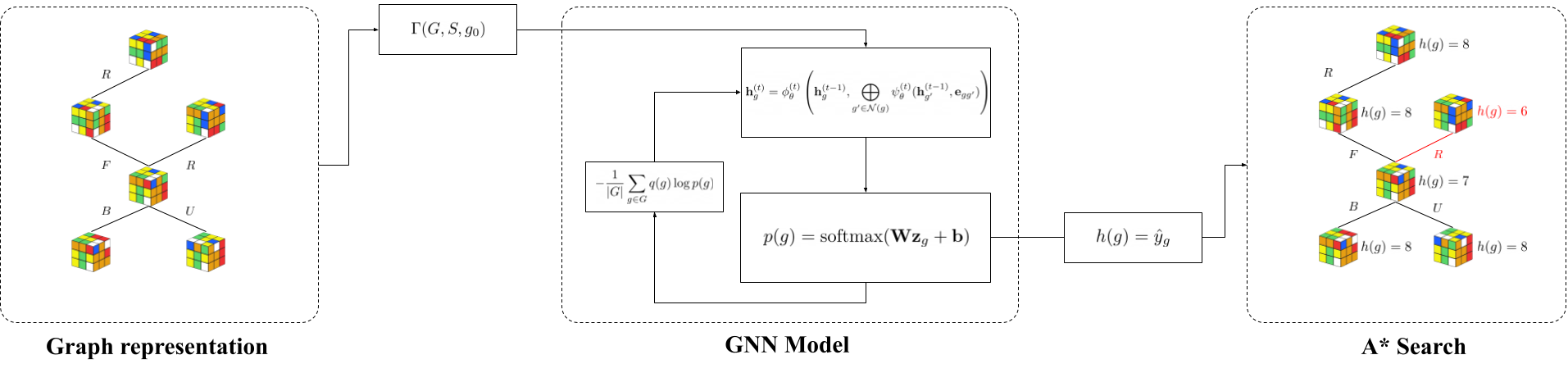}
       \caption{Method's flowchart (cube images from \url{https://rubiks-cube-solver.com/})} 
    \label{fig:enter-label}
\end{figure}

\section{Learning an Heuristic with GNNs for $A^*$ Search}
\noindent To learn $p(g)$, we employ a Graph Neural Network [\cite{Scarselli2009}], which is based on the Message Passing framework. The model requires cube states to be encoded in a feature vector $\mathbf{x}$ and iteratively aggregates information from neighboring nodes to learn embeddings. GNNs have demonstrated success in leveraging structural and feature-based relationships among adjacent nodes, particularly in classification tasks where homophily is present [\cite{Mcpherson2001}]. At each layer $t$, the hidden state $\mathbf{h}_g^{(t)}$ of a node $g$ is updated based on its previous state and messages from its neighbors
\begin{equation}
\mathbf{h}_g^{(t)} = \phi_\theta^{(t)}\left(\mathbf{h}_g^{(t-1)}, \bigoplus_{g' \in \mathcal{N}(g)} \psi_\theta^{(t)}(\mathbf{h}_{g'}^{(t-1)}, \mathbf{e}_{gg'})\right)    
\end{equation}
where $\phi_\theta^{(t)}$ and $\psi_\theta^{(t)}$ are learnable functions, $\mathcal{N}(g)$ represents the neighbors of $g$, $\mathbf{e}_{gg'}$ encodes edge features, and $\bigoplus$ is a permutation-invariant aggregation operator. After $T$ layers, the final embedding $\mathbf{z}_g=\mathbf{h}_g^{(T)}$ is used to compute the probability $p(g) = P(Y_g = y_g)$ through a classifier [\cite{Hamilton2020}]
\begin{equation}
p(g) = \mathrm{softmax}(\mathbf{W} \mathbf{z}_g + \mathbf{b})    
\end{equation}
where $\mathbf{W}$ and $\mathbf{b}$ are trainable parameters. The training data, including the subgraph $\Gamma_{\text{train}}$, the feature matrix $\mathbf{X}_{\text{train}}$, and the ground truth probabilities $q(g)$ for each training node, is collected by running $K$ random walks of length $l$ starting from the solved configuration $g_0$, with moves sampled uniformly from the available set. Additionally, by designing $l$ generator sets $S_0, S_1, \dots, S_l$ at each step of the random walk, the topology of the training subgraph (hence the quality of the gathered data) can be manipulated. However, we do not explore this aspect further, as we aim to minimize the role of human knowledge in the task.

\noindent Finally, we can utilize the label prediction $\hat{y}_g = \arg\max p(g)$ to construct a heuristic $h(g)$ for the $A^*$ search algorithm [\cite{Nilsson1968}]
\begin{equation}
    h(g) = 
    \begin{cases}
      0 & \text{if $g$ is associated with $g_0$,} \\
      \lambda \hat{y}_g & \text{otherwise}
    \end{cases}
\end{equation}
where $\lambda \in (0, \frac{1}{M}]$ is a search parameter. The upper bound on $\lambda$ ensures that the heuristic remains consistent. Specifically, if the transition cost $c(g, g') = 1$ for all $g, g' \in G$, the consistency condition $h(g) \leq c(g, g') + h(g')$ holds true for every $g, g' \in G$.

\section{Experiments}
\noindent To validate the proposed approach in practice, we compared our model against DeepCubeA on 100 test instances, which consist of cube configurations obtained by scrambling between 5 to 7 times. The reducted size of such instances is due to the limited computational power available, thus the research would highly benefit from further evaluation on larger state spaces. 

\noindent We measured key metrics such as computation time, solution path length, as well as the number of expanded nodes during the $A^*$ search. We implemented a Graph Convolutional Network (GCN) with 1 hidden layer of size 128, trained on a randomly generated graph $\Gamma_{\text{train}}$ with 100000 nodes. Such graph is the result of the execution and aggregation of 49911 random walks of length 7, where each move has a probability of $\frac{1}{12}$. As for the DeepCubeA model, we implemented a 2-layer Cost-to-Go network with hidden sizes 1000 and 200, alongside 2 residual blocks. The training phase was executed with a batch size of 100 and 1000 epochs, for a total of 100000 cube configurations.

\begin{table}[H]
    \centering
    \begin{tabular}{c|c|c|c}
     & Computation time (s) & Expanded nodes & Solution length \\
    \hline
    2-layer GCN & 372.01083 & $1.21498\times 10^5$ & 4.6930 \\
    DeepCubeA & 2.53470 & $0.90605\times 10^3$ & 5.0200 \\
    \end{tabular}
    \caption{Experiments results in terms of the studied metrics}
    \label{tab:placeholder_label}
\end{table}

\noindent Both models successfully found a solution for every test instance. The results reveal that the search phase in our model takes substantially longer than in the DeepCubeA model—approximately 146.76 times more—primarily due to the higher computational cost associated with graph convolutions, as noted in previous research [\cite{Yao2024}]. Additionally, the number of nodes expanded using the heuristic computed by the GCN is roughly 134.1 times greater than those expanded by the DAVI heuristic. Nevertheless, our model provides shorter solutions in terms of path length, about 0.33 shorter, indicating a closer approximation to optimality.

\section{Conclusion}
\noindent To summarize, we proposed a framework for a deep geometric model inspired by the DeepCubeA approach [\cite{Agostinelli2019}], aiming to minimize human intervention in the process. While our method improves solution quality by providing shorter solution paths, DeepCubeA remains the preferred choice for solving the Rubik's Cube due to its superior computational efficiency. Future research should focus on designing more representative datasets, hence optimizing the training graph structure such that group symmetries could also be leveraged. Such development could reduce the size of the training graph, enhance heuristic quality and ultimately improve the overall efficiency of the algorithm. Additionally, by extending the work to explore larger search spaces would increase validation of the proposed method's scalability and applicability. Future research will focus on the utilization of GNNs for rich state representations that can be exploited by a RL architecture. Finally, it would be valuable to test the model on other combinatorial puzzles whose structures can be represented as Cayley graphs.

\acknowledgments

\newpage

%

\end{document}